%% file: paper.tex
\documentclass[a4paper,twoside]{article}
\usepackage[dvipsnames, table]{xcolor}
 \pdfoutput=1
\usepackage{wrapfig}
\usepackage{multirow}
\usepackage{amsmath}
\usepackage{hyperref}
\usepackage{float}
\floatstyle{plaintop}
\restylefloat{table}\usepackage{epsfig}

\usepackage[T1]{fontenc}
\newcommand\thefont{\expandafter\string\the\font}

\usepackage{subcaption}
\usepackage{calc}
\usepackage{amssymb}
\usepackage{amstext}
\usepackage{amsmath}
\usepackage{amsthm}
\usepackage{multicol}
\usepackage{pslatex}
\usepackage{natbib}
\usepackage[bottom]{footmisc}
\usepackage{SCITEPRESS}

\begin{document}
\newcommand{\red}[1]{\textcolor{red}{#1}}
\definecolor{bluecol}{RGB}{20,100,255}
\thispagestyle{plain}
\pagestyle{plain}
\title{
From Xception to NEXcepTion: \\ New Design Decisions and Neural Architecture Search  }

\author{\authorname{Hadar Shavit\sup{1}, Filip Jatelnicki\sup{1}, Pol Mor-Puigventós\sup{1} and Wojtek Kowalczyk\sup{1}}
\affiliation{\sup{1}Leiden Institute of Advanced Computer Science (LIACS), Leiden University, Niels Bohrweg 1, 2333CA, The Netherlands}
\email{\{\href{mailto:h.shavit@umail.leidenuniv.nl}{h.shavit}, \href{mailto:f.o.jatelnicki@umail.leidenuniv.nl}{f.o.jatelnicki}, \href{mailto:p.mor.puigventos@umail.leidenuniv.nl}{p.mor.puigventos}\}@umail.leidenuniv.nl, \href{mailto:w.j.kowalczyk@liacs.leidenuniv.nl}{w.j.kowalczyk@liacs.leidenuniv.nl}}}


\abstract{
In this paper, we present a modified Xception architecture, the \mbox{NEXcepTion} network. Our network has significantly better performance than the original Xception, achieving \mbox{top-1} accuracy of 81.5\% on the ImageNet validation dataset (an improvement of 2.5\%) as well as a 28\% higher throughput. 
Another variant of our model, \mbox{NEXcepTion-TP}, reaches 81.8\% top-1 accuracy, similar to \mbox{ConvNeXt} (82.1\%), while having a 27\% higher throughput. 
Our model is the result of applying improved training procedures and new design decisions combined with an application of Neural Architecture Search (NAS) on a smaller dataset. These findings call for revisiting
 older architectures and reassessing their potential when combined with the latest enhancements. 
 Our code is available at \url{https://github.com/hadarshavit/NEXcepTion}.
}

\onecolumn \maketitle \normalsize \setcounter{footnote}{0} \vfill

\section{INTRODUCTION}
\label{sec:introduction}
\input{1_merged_intro}

\section{RELATED WORK}
\label{sec:related_word}
\input{2_realted_work}

\section{NEXCEPTION}
\label{sec:methods}
\input{3_methods}

\section{NEXCEPTION VARIANTS}
\label{sec:variants}
\input{4a_variants}

\section{RESULTS}
\label{sec:results}
\begin{figure*}[ht]
    \centering
    \includegraphics[width=0.75\textwidth]{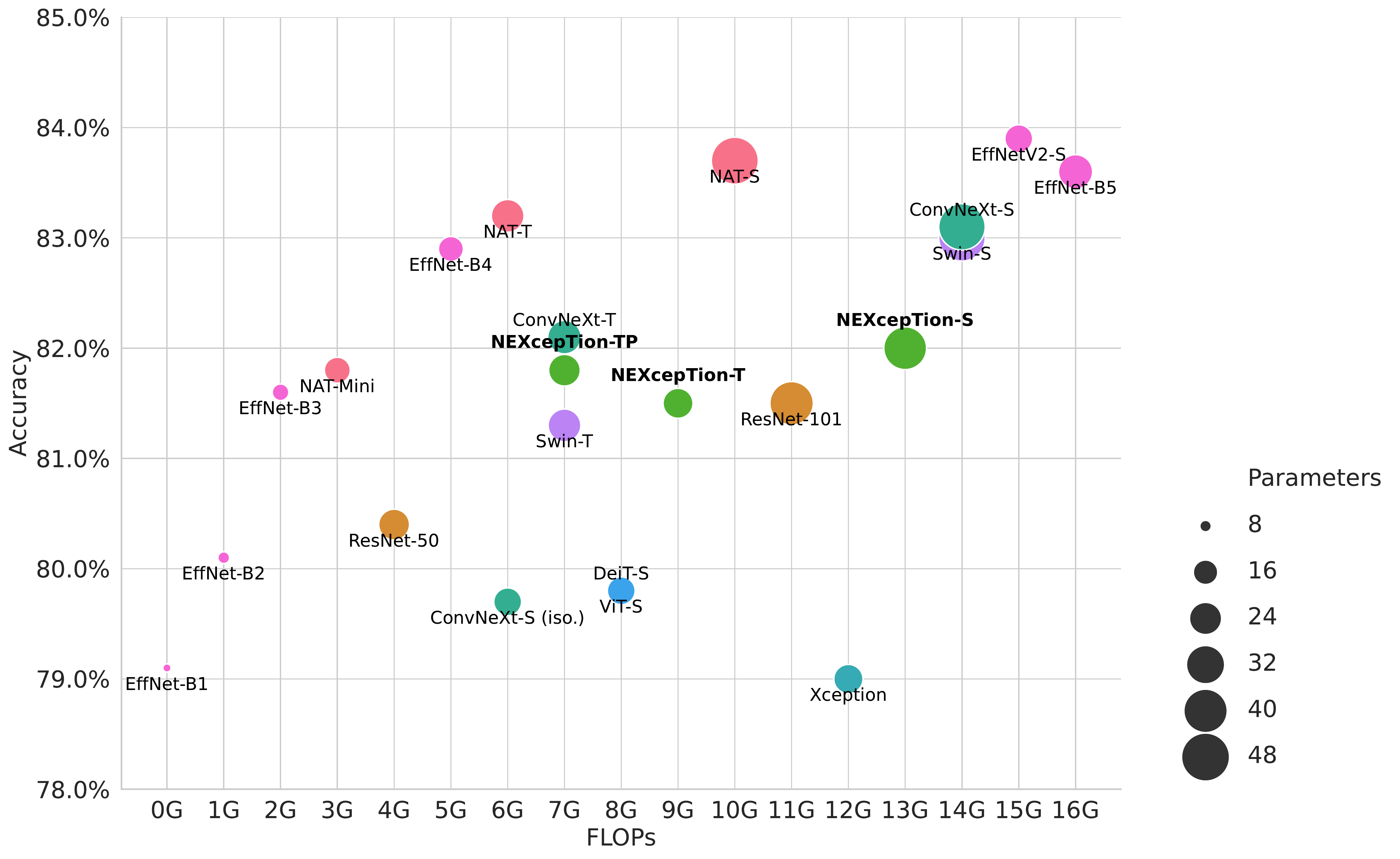}
    \caption{FLOPs and Accuracy comparison of the NEXcepTion variants (in bold), with other contemporary convolutional or Transformer networks with similar features. The size of the bubbles corresponds to the number of parameters. More details can be found in Table \ref{tab:results}, in the Appendix.}
    \label{fig:bubbles-comparison-flops}
\end{figure*}
\begin{figure*}[ht]
    \centering
    \includegraphics[width=0.75\textwidth]{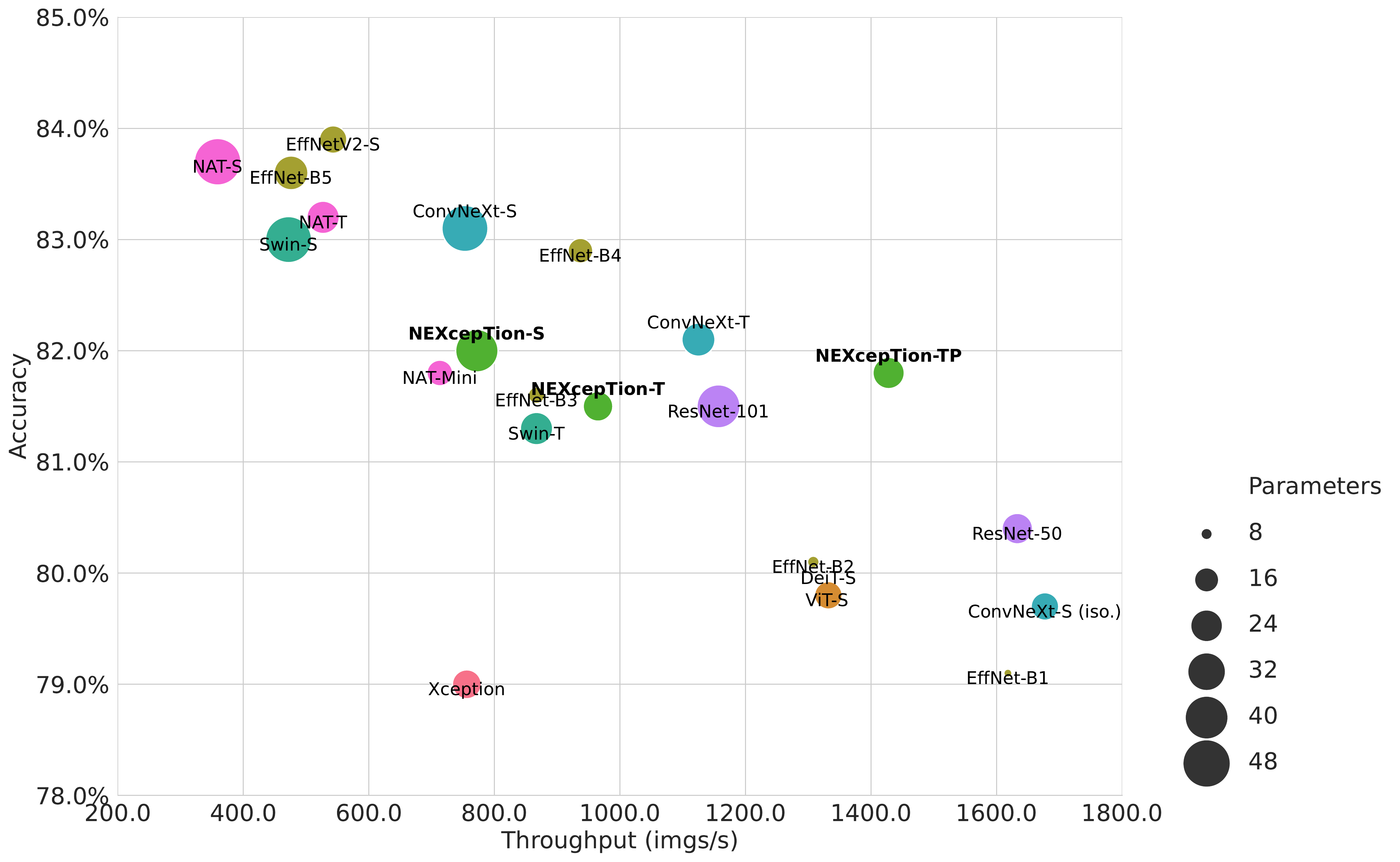}
    \caption{Throughput and Accuracy comparison of the NEXcepTion variants (in bold) with other contemporary convolutional or Transformer networks with similar features. The size of the bubbles corresponds to the number of parameters exploited. More details can be found in Table \ref{tab:results}, in the Appendix.}
    \label{fig:bubbles-comparison-throughput}
\end{figure*}

\input{4b_results}

\section{CONCLUSIONS}
\label{sec:discussion}
\input{7_discussion}

\vfill
\section*{\uppercase{ACKNOWLEDGEMENTS}}

This work was performed using the compute resources from the Academic Leiden Interdisciplinary Cluster Environment (ALICE) provided by Leiden University. We thank Andrius Bernatavicius, \mbox{Shima Javanmardi} and the participants of the Advances in Deep Learning 2022 class in LIACS for the valuable discussions and feedback.

\nocite{HendrycksEtAl2019}
\nocite{HendrycksEtAl2020}
\nocite{WangEtAl2019}
\nocite{TanLe2019}
\nocite{HasEtAl2022}
\nocite{WigEtAl2021}
\nocite{LiuEtAl2022}
\nocite{LiuEtAl2021}
\nocite{DosEtAl2020}
\nocite{TouEtAl2020}
\nocite{Chollet2016}
\nocite{TanLe2021}
\bibliographystyle{apalike}
{\small
\bibliography{paper}}

\newpage

\section*{\uppercase{Appendix}}

\input{appendix.tex}

\subsection*{Learning Curves}

\begin{figure}
    \centering
    \includegraphics[width=0.48\textwidth]{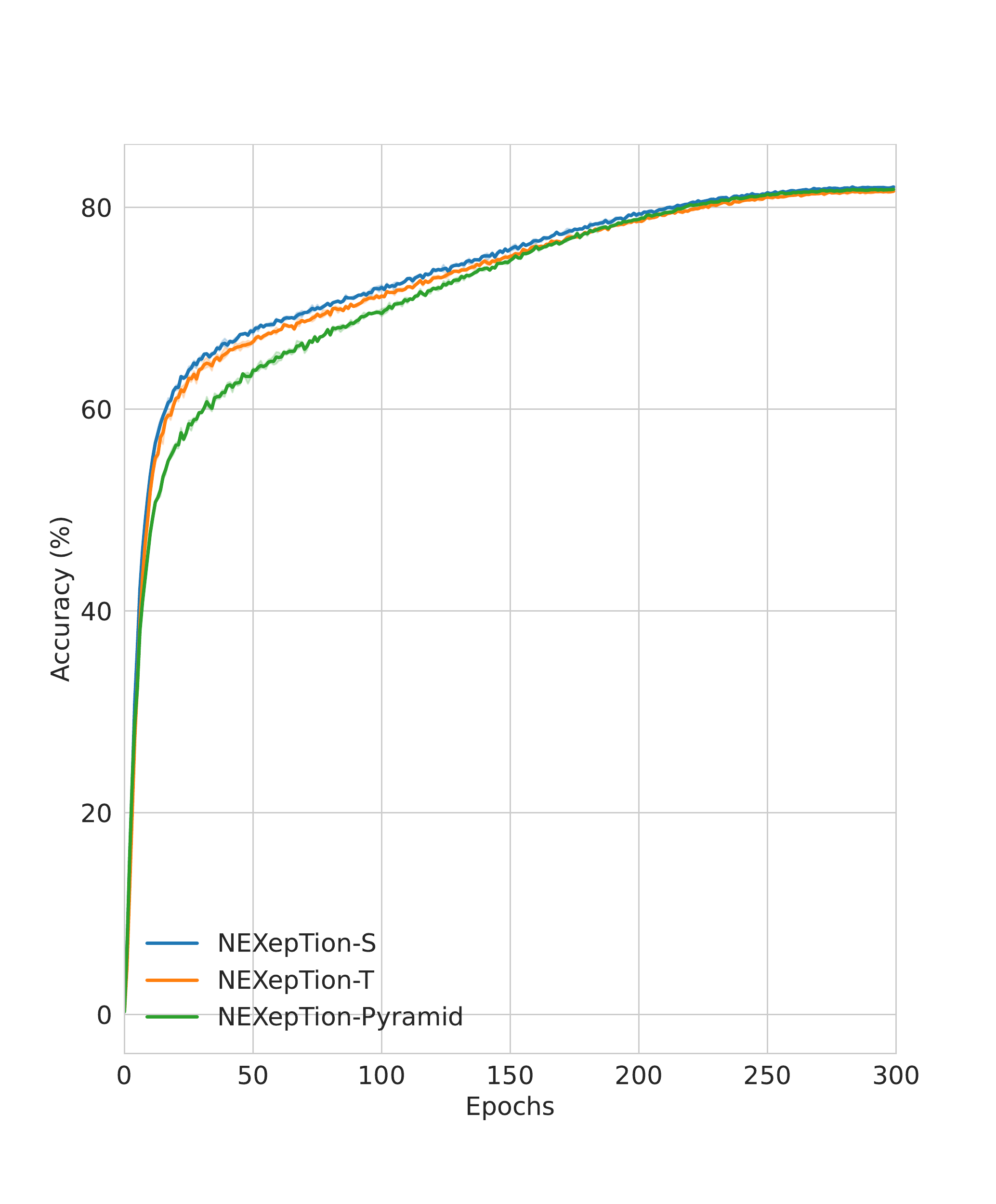}
    \caption{Accuracies on the validation set during training for all \mbox{NEXcepTion} models. Curves averaged over three random seeds with error bars.}
    \label{fig:accuracies}
\end{figure}

\begin{figure}
    \centering
    \includegraphics[width=0.48\textwidth]{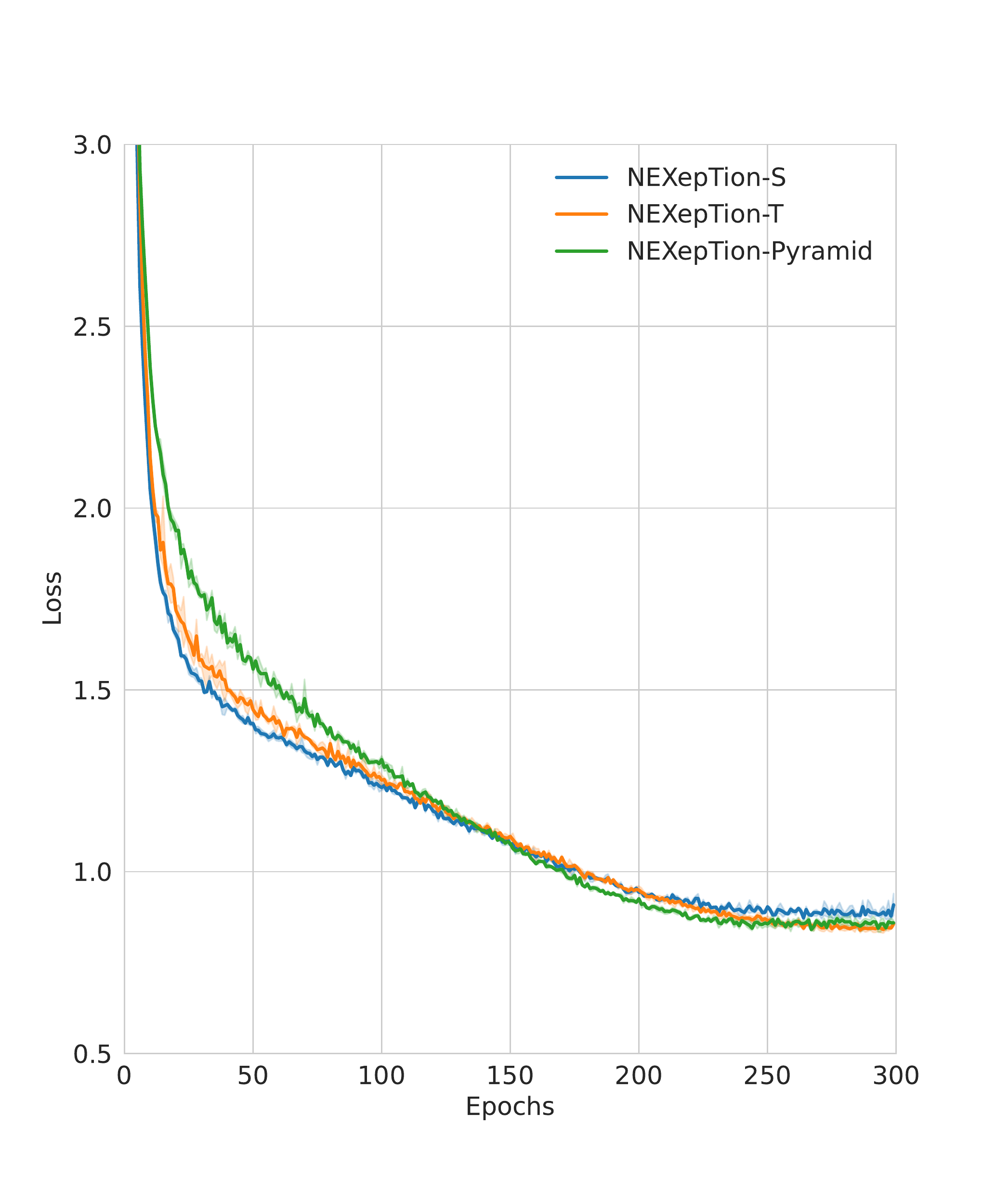}
    \caption{Cross-entropy losses on the validation set during training for all \mbox{NEXcepTion} models. Curves averaged over three random seeds with error bars.}
    \label{fig:losses}
\end{figure}

We present the accuracy and cross-entropy loss (\mbox{Figures} \ref{fig:accuracies} and \ref{fig:losses} respectively) on the ImageNet validation set during training for every  \mbox{NEXcepTion} variant. The performance of the \mbox{NEXcepTion-TP} variant is slightly delayed compared to the others while reaching a higher accuracy in the end.

\input{tab5_nexception-tp_detailed}

\input{tab2_detailed_achitectures}

\newpage
\newpage


\input{tab4_nexception-s_detailed}

\end{document}

%% file: 1_merged_intro.tex
There are multiple deep-learning-based approaches to tackle the image classification problem. In the last decade, the main attention was put into Transformers and convolutional-based architectures. Most of the recent findings in the convolutional neural networks field focused on improving the performance of the ResNet architecture \citep{LiuEtAl2022, WigEtAl2021}. In this paper we investigate how similar modifications can affect other convolutional architectures, specifically, the Xception model. In the following sections, we present \mbox{NEXcepTion}, several Xception-based models that reach state-of-the-art level accuracies. Our models are the result of running Neural Architecture Search (NAS) experiments on the CIFAR-100 dataset \citep{KrizEtAl2009} and an improved training procedure based, among other techniques, on new optimization methods and data augmentation. Our search space for experiments consists of variants of network architectures with different sizes of convolutional layers, activation functions, modern normalization and pooling methods, together with other recently introduced designs. As a final result of our experiments, we create three variants of the NEXcepTion network. All of them outperform the Xception model in the image classification task in terms of accuracy and inference throughput.
Comparing our NEXcepTion-TP model to the recently published ConvNeXt-T \citep{LiuEtAl2022}, our network reaches higher throughput ($1428\pm$9 vs. $1125\pm$5 images/second) while having a similar accuracy.

%% file: 2_realted_work.tex
\subsection{Recent Research}
During the last decade, numerous architectures have been proposed for computer vision with convolutional neural networks as their central building block.

AlexNet \citep{KriEtAl2012} trained on ImageNet with striking results, compared to the then state-of-the-art models, achieving top-1 and top-5 test set error rates of 37.5\% and 17.0\% respectively. This was the moment when state-of-the-art solutions progressed from pattern recognition to deep learning.
Thereafter, remarkable progress has been made almost yearly, starting with GoogLeNet \citep{SzegedyEtAl2014} with a more efficient (due to the \textit{Inception} module) and deeper architecture.
Later on, ResNet \citep{HeEtAl2015} was introduced, with residual (\textit{skip}) connections that allowed for even deeper networks.

In 2016, both of those last contributions were merged generating \textit{Extreme Inception}, shown in Xception \citep{Chollet2016}. This time, the \textit{Inception} module was replaced by the Xception module which used a \textit{depthwise separable convolution layer} as its basic building block.

In the following year, SENet \citep{HuEtAl2017} boosted previous networks, proposing \textit{SE-Inception} and \textit{SE-ResNet}. SENet proposed the “Squeeze-and-Excitation” (SE) block focusing on the depth dimension, recalibrating channel-wise feature responses.
EfficientNet \citep{TanLe2019} and EfficientNetV2 \citep{TanLe2021} suggested a proper scaling of existing architectures achieving superior performances.

In recent years, the machine vision community adopted the Transformers architecture, originally developed for Natural Language Processing (NLP) \citep{DevEtAl2018} by introducing the Vision Transformer (ViT) \citep{DosEtAl2020}. Since this installation of ViT, various improvements have been introduced including the Data-efficient image Transformers (DeiT) \citep{TouEtAl2020}, Swin Transformer \citep{LiuEtAl2021} and the recent Neighbourhood Attention Transformer \citep{HasEtAl2022}.

In addition to the improvements in the macro-level architectures, other micro-level improvements were introduced.
While ReLU was widely employed a few years ago, newer activation functions were published, 
for instance, the Gaussian Error Linear Unit (GELU) \citep{HenGim2016}. 

Moreover, new training procedures were exploited. While in the past basic Stochastic Gradient Descent (SGD) was used to train state-of-the-art models, today there are new variants of gradient-based optimizers such as RAdam \citep{LiuEtAl2019} AdamP \citep{HeoEtAl2020} and LAMB \citep{YouEtAl2019} with sophisticated learning rate schedules such as cosine decay \citep{LosEtAl2018}. Furthermore, data augmentation techniques such as RandAugment \citep{CubEtAl2019}, RandErase \citep{ZhongEtAl2018}, Mixup \citep{ZhaEtAl2018} and CutMix \citep{YunEtAl2019} greatly improved the accuracy of neural networks.

\subsection{Xception}

The Xception neural network was introduced by \citet{Chollet2016}. This architecture implements the \textit{depthwise separable convolution} operation. These convolutions consist of two parts: a depthwise convolution followed by a pointwise convolution. We refer to them as separable convolutional layers. The three parts of the Xception architecture are:

\textbf{Entry flow}. First, a stem of two convolutional layers of increasing sizes, followed by the first layers of the model, and then, three downsampling blocks. Each of these blocks has two separable convolutional layers with a kernel size of 3 combined with a \textit{Max Pooling} layer. Each block has a skip connection with a $1\times1$ convolution with stride 2.

\textbf{Middle flow}. The central unit contains 8 Xception blocks. Each block has three separable convolutional layers with a kernel size of 3 and stride 1. The method applied does not reshape the input size. For this reason, the size of the features map remains $19\times19\times728$ through this part of the network. In addition, there is a residual identity connection around every block.

\textbf{Exit flow}. The closure section starts with one downsampling block, like the ones in the entry flow, followed by two separable convolutions. Lastly, there is a classification head with a global average pooling and fully connected layer(s).

\subsection{Neural Architecture Search}

Neural Architecture Search (NAS) is a collection of methods for automating the design of neural network architectures \citep{ElsEtAl2019}. This can be done using automated search in a pre-defined configuration space using automated algorithm configuration methods such as Bayesian Optimization \citep{HutEtAl2011, SteEtAl2019, JinEtAl2019} or Evolutionary Algorithms \citep{LiuEtAl2020}. The usage of NAS methods has grown significantly, which can be observed in recent works like EfficientNetV2 \citep{TanLe2021}, which used NAS to improve EfficientNet.

%% file: 3_methods.tex
In this section, we present and explain our reasoning behind the chosen techniques for our search space, inspired by many recent design decisions, including ConvNeXt by \citet{LiuEtAl2022} and the re-study of ResNet by \citet{WigEtAl2021}, and extending those ideas with other innovations.

The search space is built with the PyTorch library \citep{PasEtAl2019} and \textit{timm} \citep{Wightman2019}. We apply SMAC \citep{LindauerEtAl2022} automated algorithm configurator to find a good configuration of improvements. 
Due to the considerable training time of a full network on ImageNet, we test the configurations with a reduced network, with four blocks in its main part instead of eight, and on a smaller dataset (\mbox{CIFAR-100} \citep{KrizEtAl2009})\footnote{\url{https://www.cs.toronto.edu/~kriz/cifar.html}}, with only one downsampling block in the entry flow. 
This allows trials of as many configurations as possible within 3 days on a single RTX 3090 Ti. The search space containing the combinations of parameters has more than fifty thousand different possible configurations. We optimize the architecture to \mbox{maximize accuracy.}

Our search space consists of various kernel sizes (3, 5, 7, 9), stem types (convolutional stem or patchify stem), different pooling types (max pooling, convolutional downsampling layer or blur pool),  whether to implement bottleneck in the middle flow, or to add Squeeze-and-Excitation at the end of each block. Moreover, we experiment with various positions and types of activation functions (ReLU, GELU, ELU or CELU) and with different positions and types of normalization methods (batch normalization or layer normalization). 

We performed several preliminary experiments to find the optimal training procedure. However,
we find that existing ones perform better on the final models. Therefore, we use similar training procedures to the ones created by the authors of \citet{WigEtAl2021} for the ResNet network. For more details about training procedure parameters, see Table \ref{tab:training-procedure-parameters} and the comparison between Xception and NEXcepTion architectures in Tables \ref{tab:comparison:Xception_nexception-t} and \ref{tab:comparison:Xception_nexception-s}, all in the Appendix section.

\subsection{Training Procedures}

\textbf{Stochastic Depth.}
The original Xception network performs regularization by adding a Dropout layer before the classification layer.
Stochastic depth \citep{HuangSun2016} changes the network depth during training by randomly bypassing groups of blocks and using the entire network for inference. Consequently, training time is reduced substantially and accuracy is improved introducing regularization into the network.

\textbf{Optimizer.}
In our paper, we choose \textit{Layer-wise Adaptive Moments optimizer for Batch training} (LAMB) optimizer inspired by \citet{YouEtAl2019}. As stated by \citet{WigEtAl2021}, LAMB optimizer increases the efficiency and performance of the network, in comparison to other common optimizers, like AdamW in \citet{LiuEtAl2022}, LAMB performs more accurate updates of the learning rate.

\textbf{Data Augmentation.}
While the original Xception model was trained without any data augmentation methods, newer training procedures utilize multiple techniques, which improve generalization. In our NEXcepTion model, we apply Rand Augment \citep{CubEtAl2019} that performs a few random transformations, Mixup \citep{ZhaEtAl2018} and CutMix \citep{YunEtAl2019}, which merge images, see Table \ref{tab:training-procedure-parameters} for specific values.

\textbf{Learning Rate Decay.}
Similarly to recent models such as DeiT \citep{TouEtAl2020}, we adopt cosine annealing \citep{LosEtAl2018} with warmup epochs. 
This method initially sets a low learning rate value, which gradually increases during the warmup epochs. Then, the learning rate is gradually reduced using the cosine function to achieve rapid learning. 

\subsection{Structural changes}
\textbf{``Soft'' Patchify Stem.}
Patchify layers are characterized by large kernel sizes and non-overlapping convolutions (by setting the stride and the kernel size to the same value). Inspired by this design, we add a $2\times2$ patchify layer to the search space, which we consider a ``soft'' patch, different from the aggressive $16\times16$ solution proposed by \citet{DosEtAl2020} in the Transformer schema and the $4\times4$ from ConvNeXt \citep{LiuEtAl2022}.
We use the initial block with kernel $2\times2$ and stride 2 to match the original Xception network and to fit the output size. 
This stem is adapted to the reduced resolution of the input images, similarly to the efficient configuration introduced by \citet{CorEtAl2019}.

\begin{figure}[h]
    \centering
    \includegraphics[width=0.48\textwidth]{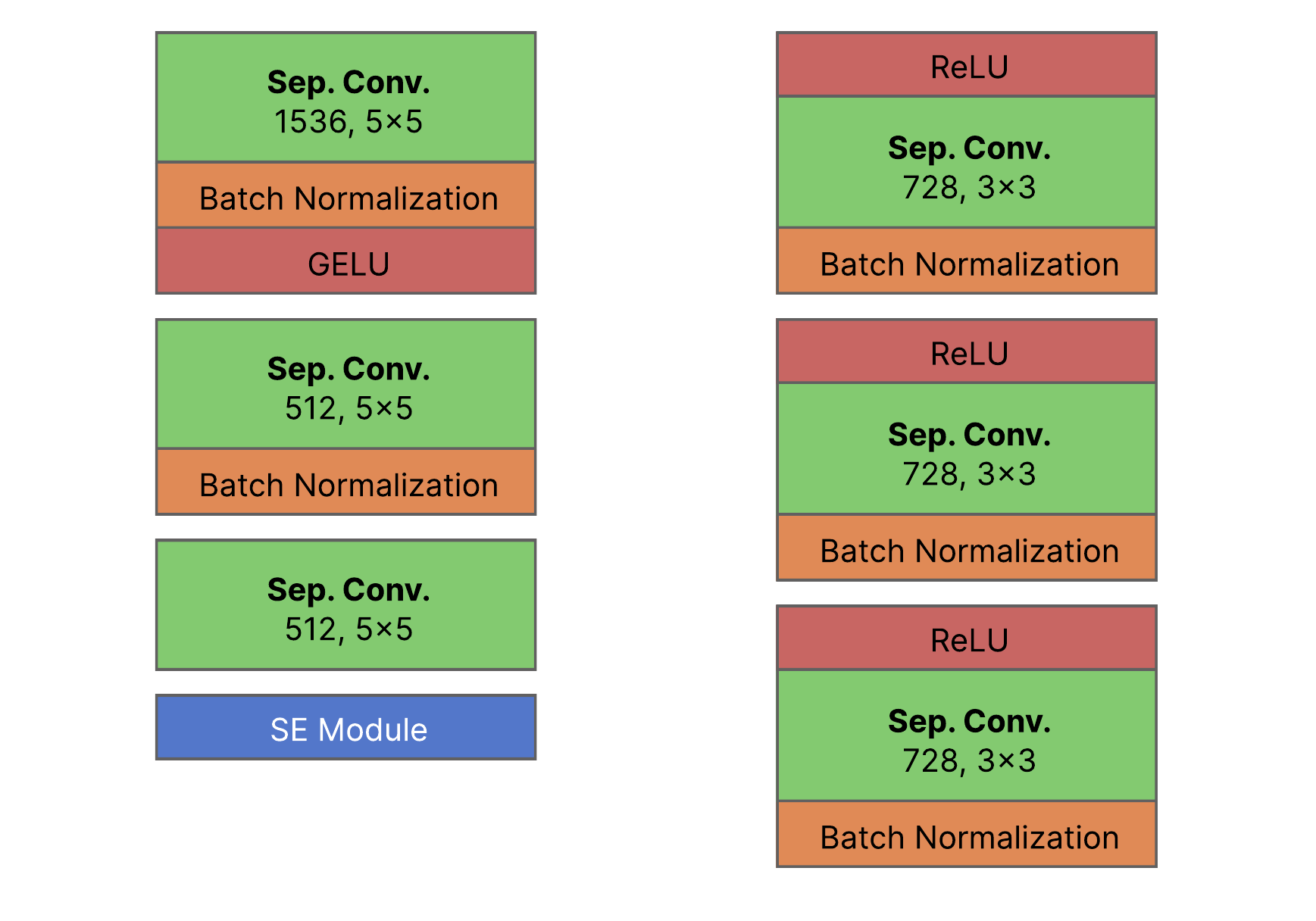}
    \caption{NEXcepTion block (left) and Xception block (right).}
    \label{fig:NEXcepTion Block and Xception Block}
\end{figure}

\textbf{Bottleneck.}
The idea of inverted bottleneck was introduced by \citet{SandlerZhu2018} and has been prevalent in modern attention-based architectures, significantly improving performance. 
The Xception architecture does not feature a bottleneck and has a constant number of channels through the middle flow of the network. In the NEXcepTion architecture, we introduce a bottleneck in the middle flow blocks, as proposed by \citet{LiuEtAl2022}, see Figure \ref{fig:NEXcepTion Block and Xception Block}.

\textbf{Larger Kernels.}
Inspired by \citet{LiuEtAl2022}, among others, we pick larger kernels for our experiments, and we achieve the best accuracy with their size set to 5. Combining this idea with bottleneck blocks and the reduced resolution allows using bigger kernels without excessive increase in the computational demand.

\newpage
\textbf{Squeeze-and-Excitation Block.}
Squeeze-and-Excitation block (SE block) from \citet{HuEtAl2017} improves channel interdependencies with an insignificant decrease in efficiency by recalibrating the feature responses channel-wise, creating superior feature maps. SE blocks provide significant performance improvement and are easy to include in existing networks as specified by \citet{HuEtAl2017}. 

\textbf{Fewer Activations and Normalizations.}

Similarly to \citet{LiuEtAl2022}, we employ less activation layers than in the original Xception network. Fewer activation layers is a distinctive property of the state-of-the-art Transformer blocks and, by replicating this concept, we can achieve higher accuracy.

Moreover, what is also inherent to Transformers architectures, is fewer normalization layers than in typical convolution-based solutions. It is important to mention that in the original Xception architecture, all convolutional layers are followed by batch normalization.

\textbf{Activation Function.} 
Concerning neuron activations, GELU \citep{HenGim2016} is used in modern Transformer architectures like BERT \citep{DevEtAl2018} and recent convolutional-based architectures like ConvNeXt \citep{LiuEtAl2022}.
Despite ReLU's simplicity and efficiency, we decide to experiment with different activation functions, inspired by the survey performed by \citet{DubeyEtAl2021}. Based on our search, we achieve the best results with the GELU activation function.

\textbf{Standardizing the Input.}
The original Xception model uses an input size of $299\times299$. We found that standardizing the input size to $224\times224$, as in \citet{HeEtAl2015}, makes the training faster on Nvidia Tensor Cores. To compensate for the lower resolution, we make the network wider by adding more channels. 

\textbf{Blur Pooling.}
Inspired by the solution from \citet{Zhang2019}, we integrate a blurring procedure before subsampling the signal. By introducing this anti-aliasing technique, our network generalizes better and achieves higher accuracy.

%% file: 4a_variants.tex
As a result of our experiments, we produce a configuration of a downsized network.

We prepare two different NEXcepTion variants, adapted to the complexities of the main ``Tiny'' and ``Small'' recent state-of-the-art models. This allows us to compare them to recent models with similar features. Additionally, we construct NEXcepTion-TP with pyramid-like architecture. All the variants use the methods described in the previous section and the NEXcepTion block presented on Figure \ref{fig:NEXcepTion Block and Xception Block}. 

\textbf{NEXcepTion-T.} 
This model exploits all the methods described in the previous section, see Table \ref{tab:comparison:Xception_nexception-t} in the Appendix. It has 24.5M parameters and 4.7 GFLOPs. The motivation for it is to have similar FLOPs and a number of parameters to the recent state-of-the-art models such as ConvNeXt \citep{LiuEtAl2022} and Swin Transformers' \citep{LiuEtAl2021} tiny models, for instance in ConvNeXt-T and Swin-T.

\textbf{NEXcepTion-S.} 
This architecture is a wider variant with 8.5 GFLOPs and 43.4M parameters. The motivation for it is to have a model with similar FLOPs to the original Xception network \citep{Chollet2016}.

\textbf{NEXcepTion-TP.} 
While the Xception architecture and the two other variants have an isotropic architecture with a constant resolution through the middle flow, other architectures such as ResNet and ConvNeXt have a pyramid-like architecture. Such architecture incorporates a few stages in its middle flow and the resolution decreases from stage to stage, hence, the name ``Pyramid''. We use the ConvNeXt architecture and replace the ConvNeXt blocks with \mbox{NEXcepTion} blocks, as well as substituting Layer Normalization with Batch Normalization and adding one more block in the second phase to have a comparable number of FLOPs. This Pyramid \mbox{NEXcepTion} model is trained with the same training procedure as  \mbox{NEXcepTion}. Our motivation is to check how the  \mbox{NEXcepTion} blocks' performance changes with the pyramid architecture, as the pyramid ConvNeXt has significantly higher accuracy than the isotropic ConvNeXt (82.1\% vs. 79.7\%). This variant has 4.5GFLOPs and 26.6M parameters.

%% file: 4b_results.tex
\subsection{DeepCAVE Analysis}

\begin{figure}[H]
    \centering
    \includegraphics[width=0.49\textwidth]{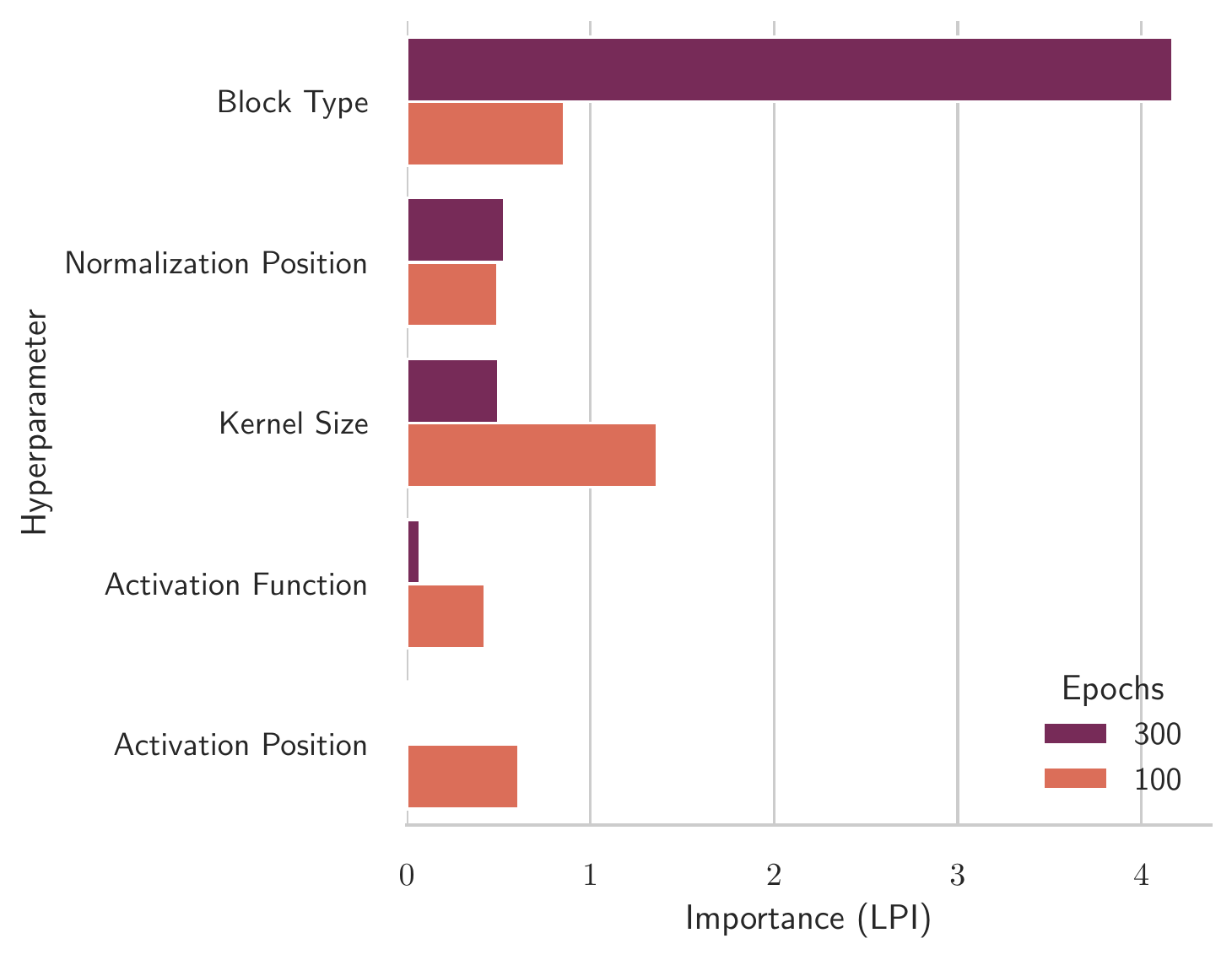}
    \caption{Local Hyperparameter Importance (LPI). Depending on the number of epochs, the influence of the selected methods differs for the final result.}
    \label{fig:importance}
\end{figure}

We first present a hyperparameter importance analysis of our NAS process, which can be seen in \mbox{Figure \ref{fig:importance}}.  We also measured the importance of the stem shape, the pooling procedure and the SE module, obtaining an importance of less than 0.1 for those features. We calculate the Local Hyperparameter Importance (LPI) using DeepCave \citep{SasEtAl2022}. We can see that the most important hyperparameter is the block type, for instance, most of the improvement comes from shifting to a bottleneck block. Changing the positions of the normalizations and the kernel sizes of the convolutions also has an impact on the performance. We can see that the activation function type has a relatively small impact on the accuracy of the model.

\subsection{Image Classification}

We train our networks on the widely used ImageNet-1K image classification benchmark\footnote{\url{https://image-net.org/}} \citep{RusEtAl2015}. We run our experiments on a single node of a local cluster. A node has 4 Nvidia RTX 2080 Ti GPUs and 2 Intel Xeon Gold 6126 2.6GHz with 12 cores CPUs. We run 3 repetitions for each of the NEXcepTion variants. Each training of the NEXcepTion-T variant takes 100 hours on average and for the NEXcepTion-TP 89 hours on average. \mbox{Finally,} the biggest model we test, the \mbox{NEXcepTion-S,} takes 150 hours on average. The top-1 accuracy of the NEXcepTion-T model is $81.6\pm0.08$, for NEXcepTion-TP is $81.7\pm0.07$, for NEXcepTion-S is $82.0\pm0.07$.

We compare our three variants with the original Xception network \citep{Chollet2016}, with the convolutional neural networks 
ConvNeXt \citep{LiuEtAl2022}, 
EfficientNetV1 \citep{TanLe2019} and 
EfficientNetV2 \citep{TanLe2021}. 
We also compare our networks with the Transformer-based models 
Vision Transformer (ViT) \citep{DosEtAl2020}, 
Data Efficient Transformer (DeiT) \citep{TouEtAl2020} and 
Swin Transformer \citep{LiuEtAl2021}. The reported accuracies of these models belong to the papers cited next to their names. We calculate the throughputs using the \textit{timm} library \citep{Wightman2019}, on a single RTX 2080 Ti with a batch size of 256, mixed precision and channels last, on the ImageNet validation dataset during 30 repetitions. For the EffNet-B4 and EffNet-B5 the calculations are made using a batch size of 128, due to GPU memory issues. 

We evaluate the models using \textit{timm}. For the NEXcepTion set models, we use our own trained weights. For isotropic \mbox{ConvNeXt} the trained weights are from \citet{LiuEtAl2022}. For Neighbourhood Attention Transformer (NAT), the trained weights are from \citet{HasEtAl2022}. The results are presented in Figures \ref{fig:bubbles-comparison-flops} and \ref{fig:bubbles-comparison-throughput}. Additionally, in the Appendix, we offer the results values in Table \ref{tab:results} and we evaluate the robustness of the NEXcepTion architectures in Table \ref{tab:robustness}.

All variants of NEXcepTion have higher accuracy than Xception, as well as higher throughput. The NEXcepTion-TP model has significantly higher throughput than the other compared models with similar accuracy.

%% file: 7_discussion.tex
In this work, we implement the backbone of the existing Xception architecture, with some modifications and improved training. We show that on the ImageNet classification task it is possible to achieve significantly higher accuracy than with the original architecture. Our findings strengthen the work published recently by \citet{LiuEtAl2022}, in which \mbox{ConvNeXt} was presented. While \mbox{ConvNeXt} only showed results on modernizing ResNet, we generalize their findings to another convolutional architecture. 

We also present a NAS method that combines both the usage of applying modern design decisions to existing architectures and automated algorithm configuration for neural architecture search. This method can be used to apply those modifications to other architectures. Nevertheless, it could be useful to generalize the usage of NAS to enhance existing architectures with other existing networks to confirm this idea.

When it comes to the obtained results, we provide three variants of the NEXcepTion network, and all reach higher accuracy and throughput than Xception. Our NEXcepTion-T outperforms the original Xception, using half of the FLOPs and a similar number of parameters. 
\newpage
In comparison to ConvNeXt, NEXcepTion-TP reaches similar accuracy with higher throughput, as reported in \mbox{Section \ref{sec:results}.} 
We note that ConvNeXt's pyramid compute ratio gives better results both in terms of accuracy and inference throughput, as using the NEXcepTion block with this compute ratio performs better than using Xception's  compute ratio. In addition, we can see that the NEXcepTion block has less impact from the compute ratio than the ConvNeXt block as the difference between the isotropic NEXcepTion-T and the pyramid NEXcepTion-TP is only 0.2 while the difference between ConvNeXt-T and isotropic ConvNeXt-S is 2.4 \citep{LiuEtAl2022}. 

Finally, we also check the generalization of our models testing their performance on robustness datasets and comparing them to other state-of-the-art models. For all datasets, the NEXcepTion set obtains better results than Xception and is frequently above the rest of the architectures, see Table \ref{tab:robustness} in the Appendix section.

Overall, this work can inspire future research to use algorithm configuration libraries like SMAC \citep{LindauerEtAl2022}, RayTune \citep{LiaEtAl2018} or KerasTuner \citep{OmaEtAl2019} as they require only the definition of a base model and a configuration space. 

Another future research direction can be to perform an in-depth importance analysis of architectural designs such as the ones we use, similarly to checking the hyperparameter's importance as in \citet{RijEtAl2018} for traditional machine learning \mbox{models}.

%% file: appendix.tex
\subsection*{Robustness Evaluation}
We check the generalization of our \mbox{NEXcepTion} models on three computer vision robustness benchmarks: ImageNet-A \citep{HendrycksEtAl2019}, ImageNet-R \citep{HendrycksEtAl2020} and ImageNet-Sketch \citep{WangEtAl2019}. We use the models trained on ImageNet-1K, without any additional data or training. A comparison of the accuracies on the robustness benchmarks is available in Table \ref{tab:robustness}. We distinguish how all the \mbox{NEXcepTion} models outperform the original Xception in all the robustness benchmarks and obtain results similar to the current state-of-the-art ConvNets and Transformers. Indeed, the variant \mbox{NEXcepTion-S} surpasses all the other compared networks on ImageNet-Sketch.

\subsection*{Architectures}

In this section we present the details of the \mbox{NEXcepTion} architectures.
In Table \ref{tab:nexception_tp} we compare ResNet-50 and \mbox{ConvNeXt-T} with our model \mbox{NEXcepTion-TP}, which is very similar to \mbox{ConvNeXt-T}. In \mbox{NEXcepTion-TP} we  replace the \mbox{ConvNeXt} block with a \mbox{NEXcepTion} block, and add one more block to match the same number of FLOPs.
In Table \ref{tab:comparison:Xception_nexception-t} we depict the differences between the original Xception architecture and the architecture of \mbox{NEXcepTion-T}. In \mbox{Table \ref{tab:comparison:Xception_nexception-s}} the differences when compared to \mbox{NEXcepTion-S} are shown. This variant is similar to \mbox{NEXcepTion}-T with more channels to match the number of FLOPs of the original Xception.

\begin{table*}[h]
\small
\centering
\rowcolors{2}{blue!0}{bluecol!7}
\begin{tabular}{@{}cccc@{}}
\hline
Training Configuration            & NEXcepTion-S       & NEXcepTion-T        & NEXcepTion-TP      \\ \hline \hline
Input Size                        & $224$  & $224$  & $224$  \\
Optimizer                         & LAMB         & LAMB         & LAMB         \\
Learning Rate                     & $1.4e^{-3}$  & $2e^{-3}$    & $2e^{-3}$    \\
Weight Decay                      & 0.02         & 0.02         & 0.02         \\
Batch Size                        & 128           & 256           & 256           \\
Training Epochs                   & 300          & 300          & 300          \\
Learning Rate Schedule            & Cosine Decay & Cosine Decay & Cosine Decay \\
Warmup Epochs                     & 5            & 5            & 5            \\
RandAugment            & 7 / 0.5    & 7 / 0.5    & 7 / 0.5            \\
Mixup                             & 0.1          & 0.1          & 0.1          \\
Cutmix                            & 1.0          & 1.0          & 1.0          \\
Random Erasing                    & 0.0         & 0.0         & 0.0       \\
Label Smoothing                   & 0            & 0            & 0            \\
Stochastic Depth                  & 0.05         & 0.05         & 0.05            \\
Minimum Learning Rate            & $1.0e^{-06}$ & $1.0e^{-06}$      & $1.0e^{-06}$      \\
Error Function & BCE & BCE & BCE\\ 
Test crop ratio                           & 0.95         & 0.95         & 0.95         \\ 
\hline
\end{tabular}
\vspace{2ex}
\caption{Training details of the proposed networks.}
\label{tab:training-procedure-parameters}
\end{table*}

\begin{table*}[h]
\centering
\rowcolors{2}{blue!0}{bluecol!7}
\small
\begin{tabular}{ccccc}
\hline
\multicolumn{1}{c}{Model} & FLOPs / Params & ImageNet-A & ImageNet-R & ImageNet-Sketch\\ \hline \hline
NEXcepTion-T    &  4.7 / 24.5                     & 17.65      & 45.99      & 34.73                               \\
NEXcepTion-TP   &  4.5 / 26.6                   & 22.75      & 47.37      & 34.70                                \\
NEXcepTion-S    &  8.5 / 43.4                    & 21.36      & 47.82      & 36.60                               \\
Xception    &      8.4 / 43.4                 & 9.83      & 40.79      & 29.88                               \\
Swin-T    &        4.5 / 28.3               & 21.60      & 41.30      & 29.10                               \\
ConvNeXt-T    &    4.5 / 28.6                   & 24.20      & 47.20      & 33.80                               \\
RVT-S    &         4.7 / 23.3              & 25.70      & 47.70      & 34.70                               \\
\hline
\end{tabular}
\caption{Robustness evaluation of NEXcepTion compared to other state-of-the-art models.}
\label{tab:robustness}
\end{table*}

\input{tab_results}

%% file: tab_results.tex
\begin{table*}[h]
\centering
\small
\rowcolors{2}{blue!0}{bluecol!7}
\scalebox{0.9}{
\begin{tabular}{lccccc}
    [$\chi/\tau$] model & \begin{tabular}[c]{@{}c@{}}input \\ img.\end{tabular} & \#params & FLOPs & 
   \begin{tabular}[c]{@{}c@{}}throughput\\ (images / s)\end{tabular} &
   \begin{tabular}[c]{@{}c@{}} IN-1K \\ top-1 acc.\end{tabular} \\
\hline \hline

$\chi$ EffNet-B1   \citep{TanLe2019} & 240$^2$ & 7.8M & 0.7G & $1618\pm24$ & 79.1 \\

$\chi$ EffNet-B2 \citep{TanLe2019} & 260$^2$ & 9.2M & 1.0G & $1308\pm9$ & 80.1 \\

$\chi$ EffNet-B3 \citep{TanLe2019} & 300$^2$ & 12M & 1.8G & $867\pm6$ & 81.6 \\

$\tau$ NAT-Mini \citep{HasEtAl2022} & 224$^2$ & 20M & 2.7G & $713\pm3$ & 81.8 \\

\hline
$\chi$ ResNet-50 \citep{WigEtAl2021} & 224$^2$ & 25.6M & 4.1G & $1633\pm26$ & 80.4\\

$\chi$ EffNet-B4 \citep{TanLe2019} & 380$^2$ & 19M & 4.2G & $937\pm6$ & 82.9 \\

$\chi$ ConvNeXt-S (iso.) \citep{LiuEtAl2022} & 224$^2$ & 22M & 4.3G & $1677\pm77$ & 79.7 \\ 

$\tau$ NAT-T \citep{HasEtAl2022} & 224$^2$ & 28M & 4.3G & $527\pm1$ & 83.2 \\

$\tau$ Swin-T \citep{LiuEtAl2021}  & 224$^2$ & 28M & 4.5G & $867\pm7$ & 81.3 \\

\rowcolor{Goldenrod!30}
$\chi$ NEXcepTion-TP & 224$^2$ & 26.6M &  4.5G & 1428 $\pm$ 9 & 81.8\\

$\chi$ ConvNeXt-T \citep{LiuEtAl2022} & 224$^2$ & 29M & 4.5G & 1125 $\pm5$ & 82.1 \\

$\tau$ ViT-S \citep{DosEtAl2020, LiuEtAl2022} & 224$^2$ & 22M & 4.6G &  $1330\pm8$ & 79.8\\

$\tau$ DeiT-S \citep{TouEtAl2020} & 224$^2$ & 22M & 4.6G & $1332\pm13$ & 79.8 \\ 

\rowcolor{Goldenrod!30}
$\chi$ NEXcepTion-T & 224$^2$ & 24.5M & 4.7G  & $965\pm6$ & 81.5\\

\hline

$\tau$ NAT-S \citep{HasEtAl2022} & 224$^2$ & 51M & 7.8G & $359\pm1$ & 83.7 \\

$\chi$ ResNet-101 \citep{WigEtAl2021} & 224$^2$ & 44.5M & 7.9G & $1157\pm6$ & 81.5\\

$\chi$ Xception \citep{Chollet2016} & 299$^2$ & 23.6M & 8.4G & $756\pm5$ & 79.0 \\

\rowcolor{Goldenrod!30}
$\chi$ NEXcepTion-S & 224$^2$ & 43.4M & 8.5G  & $772\pm3$ & 82.0\\

$\tau$ Swin-S \citep{LiuEtAl2021} & 224$^2$ & 50M & 8.7G & $472\pm1$ & 83.0 \\

$\chi$ ConvNeXt-S \citep{LiuEtAl2022} & 224$^2$ & 50M & 8.7G & $753\pm3$ & 83.1 \\ 

$\chi$ EffNetV2-S \citep{TanLe2021} & 384$^2$ & 22M & 8.8G & $543\pm6$ & 83.9 \\

$\chi$ EffNet-B5 \citep{TanLe2019} & 456$^2$  & 30M & 9.9G & $476\pm2$ & 83.6 \\

\hline

\end{tabular}}
\caption{Classification accuracy on ImageNet-1K.  $\chi$ for Convolutional and $\tau$ for Transformer networks. We present the results sorted by their FLOPs value. The values are obtained from the cited publications, except for the throughput's, calculated using the \textit{timm} library on a RTX 2080 Ti during 30 repetitions.}
\label{tab:results}
\vspace{-1.5em}
\end{table*}

%% file: tab5_nexception-tp_detailed.tex
\begin{table*}[t]
\small
\centering
\addtolength{\tabcolsep}{-2pt}
\vspace{2ex}
{
\begin{tabular}{c|c|c|c|c}
\small
 &  output size &  ResNet-50 & ConvNeXt-T  &  NeXcepTion-TP \\\hline
\multirow{2}{*}{stem} & \multirow{2}{*}{\begin{tabular}[c]{@{}c@{}} 56$\times$56 \end{tabular}} & 
7$\times$7, 64, stride 2  
& \multirow{2}{*}{4$\times$4, 96, stride 4} & \multirow{2}{*}{4$\times$4, 96, stride 4} \\ 

& & 3$\times$3 max pool, stride 2 & & \\

\hline
\multirow{5}{*}{res2} & 
\multirow{5}{*}{\begin{tabular}[c]{@{}c@{}} 56$\times$56 \end{tabular}} & 
\multirow{5}{*}{$\begin{bmatrix}\text{1$\times$1, 64}\\\text{3$\times$3, 64}\\\text{1$\times$1, 256}\end{bmatrix}$ $\times$ 3}  & 
\multirow{5}{*}{$\begin{bmatrix}\text{d7$\times$7, 96}\\\text{1$\times$1, 384}\\\text{1$\times$1, 96}\end{bmatrix}$ $\times$ 3} &
\multirow{5}{*}{$\begin{bmatrix}\text{d5$\times$5, 288}\\\text{d5$\times$5, 96}\\\text{d5$\times$5, 96}\\\text{SE Module}\end{bmatrix}$ $\times$ 3} \\
& & & & \\
& & & & \\
& & & & \\
& & & & \\
\hline
\multirow{5}{*}{res3} & 
\multirow{5}{*}{\begin{tabular}[c]{@{}c@{}} 28$\times$28  \end{tabular}} & 
\multirow{5}{*}{$\begin{bmatrix}\text{1$\times$1, 128}\\\text{3$\times$3, 128}\\\text{1$\times$1, 512}\end{bmatrix}$ $\times$ 4}  & 
\multirow{5}{*}{$\begin{bmatrix}\text{d7$\times$7, 192}\\\text{1$\times$1, 768}\\\text{1$\times$1, 192}\end{bmatrix}$ $\times$ 3} &
\multirow{5}{*}{$\begin{bmatrix}\text{d5$\times$5, 576}\\\text{d5$\times$5, 192}\\\text{d5$\times$5, 192}\\\text{SE Module}\end{bmatrix}$ $\times$ 4} \\
& & & & \\
& & & & \\
& & & & \\
& & & & \\
\hline
\multirow{5}{*}{res4} & 
\multirow{5}{*}{\begin{tabular}[c]{@{}c@{}} 14$\times$14 \end{tabular}} & 
\multirow{5}{*}{$\begin{bmatrix}\text{1$\times$1, 256}\\\text{3$\times$3, 256}\\\text{1$\times$1, 1024}\end{bmatrix}$ $\times$ 6}  & 
\multirow{5}{*}{$\begin{bmatrix}\text{d7$\times$7, 384}\\\text{1$\times$1, 1536}\\\text{1$\times$1, 384}\end{bmatrix}$ $\times$ 9} &
\multirow{5}{*}{$\begin{bmatrix}\text{d5$\times$5, 1152}\\\text{d5$\times$5, 384}\\\text{d5$\times$5, 384}\\\text{SE Module}\end{bmatrix}$ $\times$ 9} \\
& & & & \\
& & & & \\
& & & & \\
& & & & \\
\hline
\multirow{5}{*}{res5} & 
\multirow{5}{*}{\begin{tabular}[c]{@{}c@{}} 7$\times$7 \end{tabular}} & 
\multirow{5}{*}{$\begin{bmatrix}\text{1$\times$1, 512}\\\text{3$\times$3, 512}\\\text{1$\times$1, 2048}\end{bmatrix}$ $\times$ 3}  & 
\multirow{5}{*}{$\begin{bmatrix}\text{d7$\times$7, 768}\\\text{1$\times$1, 3072}\\\text{1$\times$1, 768}\end{bmatrix}$ $\times$ 3} &
\multirow{5}{*}{$\begin{bmatrix}\text{d5$\times$5, 2304}\\\text{d5$\times$5, 768}\\\text{d5$\times$5, 768}\\\text{SE Module}\end{bmatrix}$ $\times$ 3} \\
& & & & \\
& & & & \\
& & & & \\
& & & & \\
\hline
\multicolumn{2}{c|}{FLOPs}
&
$4.1 \times 10^9$
&
$4.5 \times 10^9$
&
$4.5 \times 10^9$
\\
\hline
\multicolumn{2}{c|}{\# params.}
&
$25.6 \times 10^6$
&
$28.6 \times 10^6$
&
$56.6 \times 10^6$ \\
\hline

\end{tabular}
}
\normalsize
\caption{Detailed architecture specifications for ResNet-50, ConvNeXt-T and NEXcepTion-TP.}
\label{tab:nexception_tp}
\end{table*}

%% file: tab2_detailed_achitectures.tex
\begin{table*}[t]
\small
\centering
{
\begin{tabular}{|c|cl|cl|}
\hline
 &
  \multicolumn{2}{c|}{Xception} &
  \multicolumn{2}{c|}{NEXcepTion-T} \\ \hline
Input size &
  $299\times299\times3$&
   &
  $224\times224\times3$ &
   \\ \hline
\multirow{4}{*}{Stem} &
  \multicolumn{1}{c}{} &
   &
  \multicolumn{1}{c}{} &
   \\
 &
  $3\times3$, 32, $s=2$ &
   &
  \multirow{2}{*}{$2\times2$, 96, $s=2$} &
   \\
 &
  $3\times3$, 64 &
   &
   &
   \\
 &
  \multicolumn{1}{c}{} &
   &
  \multicolumn{1}{c}{} &
   \\ \hline
\multirow{15}{*}{Entry flow} &
  \multicolumn{1}{c}{} &
   &
  \multicolumn{1}{c}{} &
   \\
 &
  $3\times3$, 128 &
  \multicolumn{1}{c|}{|} &
  $5\times5$, 128 &
  \multicolumn{1}{c|}{|} \\
 &
  $3\times3$, 128 &
  \multicolumn{1}{c|}{{[}$1\times1$, s=2{]}} &
  $5\times5$, 128 &
  \multicolumn{1}{c|}{{[}$1\times1$, s=2{]}} \\
 &
  MaxPool $3\times3$ &
  \multicolumn{1}{c|}{|} &
  MaxBlurPool $3\times3$ &
  \multicolumn{1}{c|}{|} \\
 & 
  \multicolumn{1}{c}{} &
   & SE Module &
  \multicolumn{1}{c|}{|} 
   \\
 &
  \multicolumn{1}{c}{} &
  &
  \multicolumn{1}{c}{} &
  \\
 &
  $3\times3$, 256 &
  \multicolumn{1}{c|}{|} &
  $5\times5$, 256 &
  \multicolumn{1}{c|}{|} \\
 &
  $3\times3$, 256 &
  \multicolumn{1}{c|}{{[}$1\times1$, s=2{]}} &
  $5\times5$, 256 &
  \multicolumn{1}{c|}{{[}$1\times1$, s=2{]}} \\
 &
  MaxPool $3\times3$ &
  \multicolumn{1}{c|}{|} &
  MaxBlurPool $3\times3$ &
  \multicolumn{1}{c|}{|} \\
 &
  \multicolumn{1}{c}{} &
   & SE Module &
  \multicolumn{1}{c|}{|} 
   \\
 &
  \multicolumn{1}{c}{} 
  &
  & 
  \multicolumn{1}{c}{} &
  \\
 &
  $3\times3$, 728 &
  \multicolumn{1}{c|}{|} &
  $5\times5$, 512 &
  \multicolumn{1}{c|}{|} \\
 &
  $3\times3$, 728 &
  \multicolumn{1}{c|}{{[}$1\times1$, s=2{]}} &
  $5\times5$, 512 &
  \multicolumn{1}{c|}{{[}$1\times1$, s=2{]}} \\
 &
  MaxPool $3\times3$ &
  \multicolumn{1}{c|}{|} &
  MaxBlurPool $3\times3$ &
  \multicolumn{1}{c|}{|} \\
 &
  \multicolumn{1}{c}{} &
   & SE Module &
  \multicolumn{1}{c|}{|} 
   \\ \hline
\begin{tabular}[c]{@{}l@{}}Resulting \\ feature maps\end{tabular} &
  $19\times19\times728$ &
   &
  $14\times14\times512$ &
   \\ \hline
\multirow{5}{*}{\begin{tabular}[c]{@{}l@{}}Middle flow \\ \\ $\times \ 8$\end{tabular}} &
  \multicolumn{1}{c}{} &
   &
  \multicolumn{1}{c}{} &
   \\
 &
  $3\times3$, 728 &
  \multicolumn{1}{c|}{|} &
  $5\times5$, 1536 &
  \multicolumn{1}{c|}{|} \\
 &
  $3\times3$, 728 &
  \multicolumn{1}{c|}{$\times1$} &
  $5\times5$, 512 &
  \multicolumn{1}{c|}{$\times1$} \\
 &
  $3\times3$, 728 &
  \multicolumn{1}{c|}{|} &
  $5\times5$, 512 &
  \multicolumn{1}{c|}{|} \\
 &
  \multicolumn{1}{c}{} &
   & SE Module &
  \multicolumn{1}{c|}{|} 
   \\ \hline
\begin{tabular}[c]{@{}l@{}}Resulting \\ feature maps\end{tabular} &
  $19\times19\times728$ &
   &
  $14\times14\times512$ &
   \\ \hline
\multirow{9}{*}{Exit flow} &
  \multicolumn{1}{c}{} &
   &
  \multicolumn{1}{c}{} &
   \\
 &
  $3\times3$, 728 &
  \multicolumn{1}{c|}{|} &
  $5\times5$, 512 &
  \multicolumn{1}{c|}{|} \\
 &
  $3\times3$, 1024 &
  \multicolumn{1}{c|}{{[}$1\times1$, s=2{]}} &
  $5\times5$, 1024 &
  \multicolumn{1}{c|}{{[}$1\times1$, s=2{]}} \\
 &
  MaxPool $3\times3$ &
  \multicolumn{1}{c|}{|} &
  MaxBlurPool $3\times3$ &
  \multicolumn{1}{c|}{|} \\
 &
  \multicolumn{1}{c}{} &
   & SE Module &
  \multicolumn{1}{c|}{|} 
  
  \\
 &
  \multicolumn{1}{c}{} &
   &
  \multicolumn{1}{c}{} &
  
   \\
 &
  $3\times3$, 1536 &
   &
  $3\times3$, 1536 &
   \\
 &
  $3\times3$, 2048 &
   &
  $3\times3$, 2048  &
   \\
 &
  Global Average Pooling &
   &
  Global Average Pooling &
   \\
 &
  \multicolumn{1}{c}{} &
   &
  \multicolumn{1}{c}{} &
   \\ \hline
Output &
  \begin{tabular}[c]{@{}c@{}}Fully connected layers \end{tabular} &
   &
  \begin{tabular}[c]{@{}c@{}}Fully connected layers \end{tabular} &
   \\ \hline
\end{tabular}
}
\centering
\vspace{1ex}
\caption{Architecture comparison between Xception and NEXcepTion-T. In the right side of each model the residual layers are described for each block. All the convolutions are depthwise separable convolutions except for the convolutions on the stem block and on the residual connections. The MaxPool operations have stride 2.}
\label{tab:comparison:Xception_nexception-t}
\end{table*}

%% file: tab4_nexception-s_detailed.tex
\begin{table*}[t]
\centering
\small

{
\begin{tabular}{|l|cl|cl|}
\hline
 &
  \multicolumn{2}{c|}{Xception} &
  \multicolumn{2}{c|}{NEXcepTion-S} \\ \hline
Input size &
  $299\times299\times3$ &
   &
  $224\times224\times3$ &
   \\ \hline
\multirow{4}{*}{Stem} &
  \multicolumn{1}{l}{} &
   &
  \multicolumn{1}{l}{} &
   \\
 &
  $3\times3$, 32, $s=2$ &
   &
  \multirow{2}{*}{$2\times2$, 96, $s=2$} &
   \\
 &
  $3\times3$, 64 &
   &
   &
   \\
 &
  \multicolumn{1}{l}{} &
   &
  \multicolumn{1}{l}{} &
   \\ \hline
\multirow{15}{*}{Entry flow} &
  \multicolumn{1}{l}{} &
   &
  \multicolumn{1}{l}{} &
   \\
 &
  $3\times3$, 128 &
  \multicolumn{1}{c|}{|} &
  $5\times5$, 128 &
  \multicolumn{1}{c|}{|} \\
 &
  $3\times3$, 128 &
  \multicolumn{1}{c|}{{[}$1\times1$, s=2{]}} &
  $5\times5$, 128 &
  \multicolumn{1}{c|}{{[}$1\times1$, s=2{]}} \\
 &
  MaxPool $3\times3$ &
  \multicolumn{1}{c|}{|} &
  MaxBlurPool $3\times3$ &
  \multicolumn{1}{c|}{|} \\
 &
  \multicolumn{1}{l}{} &
   & SE Module &
  \multicolumn{1}{c|}{|} 
   \\
 &
  \multicolumn{1}{l}{} &
   &
  \multicolumn{1}{l}{} &
   \\
 &
  $3\times3$, 256 &
  \multicolumn{1}{c|}{|} &
  $5\times5$, 256 &
  \multicolumn{1}{c|}{|} \\
 &
  $3\times3$, 256 &
  \multicolumn{1}{c|}{{[}$1\times1$, s=2{]}} &
  $5\times5$, 256 &
  \multicolumn{1}{c|}{{[}$1\times1$, s=2{]}} \\
 &
  MaxPool $3\times3$ &
  \multicolumn{1}{c|}{|} &
  MaxBlurPool $3\times3$ &
  \multicolumn{1}{c|}{|} \\
 &
  \multicolumn{1}{l}{} &
   & SE Module &
  \multicolumn{1}{c|}{|} 
   \\
 &
  \multicolumn{1}{l}{} &
   &
  \multicolumn{1}{l}{} &
   \\
 &
  $3\times3$, 728 &
  \multicolumn{1}{c|}{|} &
  $5\times5$, 752 &
  \multicolumn{1}{c|}{|} \\
 &
  $3\times3$, 728 &
  \multicolumn{1}{c|}{{[}$1\times1$, s=2{]}} &
  $5\times5$, 752 &
  \multicolumn{1}{c|}{{[}$1\times1$, s=2{]}} \\
 &
  MaxPool $3\times3$ &
  \multicolumn{1}{c|}{|} &
  MaxBlurPool $3\times3$ &
  \multicolumn{1}{c|}{|} \\
 &
  \multicolumn{1}{l}{} &
   & SE Module &
  \multicolumn{1}{c|}{|} 
   \\ \hline
\begin{tabular}[c]{@{}l@{}}Resulting \\ feature maps\end{tabular} &
  $19\times19\times728$ &
   &
  $14\times14\times752$ &
   \\ \hline
\multirow{5}{*}{\begin{tabular}[c]{@{}l@{}}Middle flow \\ \\ $\times \ 8$\end{tabular}} &
  \multicolumn{1}{l}{} &
   &
  \multicolumn{1}{l}{} &
   \\
 &
  $3\times3$, 728 &
  \multicolumn{1}{c|}{|} &
  $5\times5$, 2256 &
  \multicolumn{1}{c|}{|} \\
 &
  $3\times3$, 728 &
  \multicolumn{1}{c|}{$\times1$} &
  $5\times5$, 752 &
  \multicolumn{1}{c|}{$\times1$} \\
 &
  $3\times3$, 728 &
  \multicolumn{1}{c|}{|} &
  $5\times5$, 752 &
  \multicolumn{1}{c|}{|} \\
 &
  \multicolumn{1}{l}{} &
   & SE Module &
  \multicolumn{1}{c|}{|} 
   \\ \hline
\begin{tabular}[c]{@{}l@{}}Resulting \\ feature maps\end{tabular} &
  $19\times19\times728$ &
   &
  $14\times14\times752$ &
   \\ \hline
\multirow{9}{*}{Exit flow} &
  \multicolumn{1}{l}{} &
   &
  \multicolumn{1}{l}{} &
   \\
 &
  $3\times3$, 728 &
  \multicolumn{1}{c|}{|} &
  $5\times5$, 752 &
  \multicolumn{1}{c|}{|} \\
 &
  $3\times3$, 1024 &
  \multicolumn{1}{c|}{{[}$1\times1$, s=2{]}} &
  $5\times5$, 1024 &
  \multicolumn{1}{c|}{{[}$1\times1$, s=2{]}} \\
 &
  MaxPool $3\times3$ &
  \multicolumn{1}{c|}{|} &
  MaxBlurPool $3\times3$ &
  \multicolumn{1}{c|}{|} \\
 &
  \multicolumn{1}{l}{} &
   & SE Module &
  \multicolumn{1}{c|}{|} 
   \\

 &
  \multicolumn{1}{c}{} &
   &
  \multicolumn{1}{c}{} &
  
   \\
 &
  $3\times3$, 1536 &
   &
  $3\times3$, 1536 &
   \\
 &
  $3\times3$, 2048 &
   &
  $3\times3$, 2048 &
   \\
 &
  Global Average Pooling &
   &
  Global Average Pooling &
   \\
 &
  \multicolumn{1}{l}{} &
   &
  \multicolumn{1}{l}{} &
   \\ \hline
Output &
  \begin{tabular}[c]{@{}c@{}}Fully connected layers\end{tabular} &
   &
  \begin{tabular}[c]{@{}c@{}}Fully connected layers\end{tabular} &
   \\ \hline
\end{tabular}
}
\centering
\vspace{1ex}
\caption{Architecture comparison between Xception and NEXcepTion-S. In the right side of each model the residual layers are described for each block. All the convolutions are \textit{depthwise separable convolutions} except for the convolutions on the stem block and on the residual connections. The MaxPool operations have stride 2.}
\label{tab:comparison:Xception_nexception-s}
\end{table*}